\documentclass[sigconf, nonacm]{acmart}

\AtBeginDocument{%
  \providecommand\BibTeX{{%
    \normalfont B\kern-0.5em{\scshape i\kern-0.25em b}\kern-0.8em\TeX}}}

\acmSubmissionID{2430}

\usepackage{footnote}
\usepackage{multirow}
\usepackage{booktabs}
\usepackage{threeparttable}
\usepackage{enumerate}
\usepackage{indentfirst}
\usepackage{array}
\usepackage{rotating}
\usepackage{ragged2e}
\usepackage{wasysym}
\usepackage{caption}

\def\etal{{\em et al.}}

\newcommand{\figref}[1]{Fig. \ref{#1}}
\newcommand{\tabref}[1]{Tab. \ref{#1}}

\newcommand{\myPara}[1]{\vspace{.05in}\noindent\textbf{#1}}

\newcommand{\bl}[1]{\textbf{#1}}
\newcommand{\mc}[1]{\mathcal{#1}}
\newcommand{\mb}[1]{\mathbb{#1}}

\begin{document}

\title{Intrinsic Relationship Reasoning for Small Object Detection}


\author{Kui Fu$^{1, 4}$, Jia Li$^{1, 4}$, Lin Ma$^{2}$, Kai Mu$^{1}$ and Yonghong Tian$^{3, 4}$}
\authornote{Corresponding author: Jia Li (website: http://cvteam.net).}
\affiliation{%
  \institution{$^1$State Key Lab. of Virtual Reality Technology and Systems, Beihang University}
  \institution{$^2$Tencent AI Laboratory, Shenzhen, China}
  \institution{$^3$School of Electronics Engineering and Computer Science, Peking University}
  \institution{$^4$Peng Cheng Laboratory, Shenzhen, China}
}
\renewcommand{\shortauthors}{Fu, et al.}

\begin{abstract}
  The small objects in images and videos are usually not independent individuals. Instead, they more or less present some semantic and spatial layout relationships with each other. Modeling and inferring such intrinsic relationships can thereby be beneficial for small object detection. In this paper, we propose a novel context reasoning approach for small object detection which models and infers the intrinsic semantic and spatial layout relationships between objects. Specifically, we first construct a semantic module to model the sparse semantic relationships based on the initial regional features, and a spatial layout module to model the sparse spatial layout relationships based on their position and shape information, respectively. Both of them are then fed into a context reasoning module for integrating the contextual information with respect to the objects and their relationships, which is further fused with the original regional visual features for classification and regression. Experimental results reveal that the proposed approach can effectively boost the small object detection performance.

\end{abstract}




\maketitle
\section{Introduction}
In recent years, various object detection approaches have boomed, which can attribute to the great success of deep convolutional neural networks (CNNs) \cite{girshick2014rich,ren2015faster,liu2016ssd}. However, the performance of the majority of CNN-based detectors \cite{he2017mask,redmon2016you} for the small objects is still far from satisfactory since they extract semantically strong features via stacking deep convolutional neural layers, which is usually accompanied with non-negligible spatial information attenuation. Therefore, a crucial challenge for small object detection is how to capture semantically strong features and simultaneously minimize spatial information attenuation.

\begin{figure}[t]
\begin{center}
   \includegraphics[width=1.0\columnwidth]{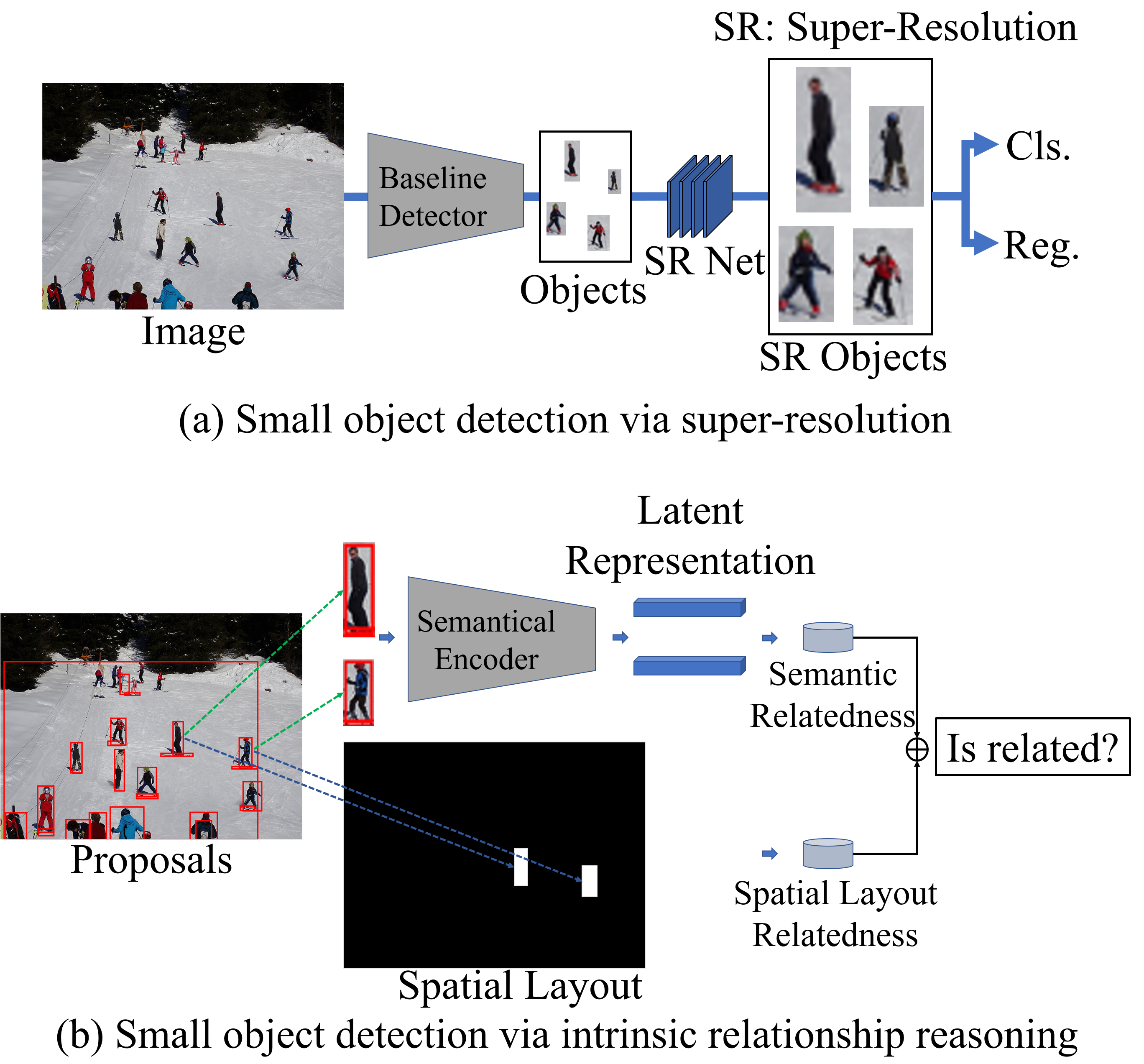}
\end{center}
   \caption{Comparison of different strategies for small object detection: a) Using a super-resolution network to up-sample a blurry low-resolution image, where one baseline detector is performed, to fine-scale high-resolution one, on which the object detection results are refined. b) Our proposed intrinsic relationship graph construction. The region-to-region object pair is fed into the semantic encoder and calculated semantic relatedness. Simultaneously, the spatial layout is exploited to calculate the spatial relatedness. Then the intrinsic relationship can be well modeled through integrating the semantic and spatial layout relatedness. Note that the two strategies can complementary to each other.}
\label{fig:motivation}
\end{figure}

There is an increasing concern about small object detection. Bai \etal~\cite{bai2018finding,bai2018sod} proposes an intuitive and effective solution, as illustrated in \figref{fig:motivation} (a), which employs a super-resolution network to up-sample a blurry low-resolution image to fine-scale high-resolution one, on which the detection results are refined. Such an approach fundamentally solves the spatial information attenuation problem, but at the cost of the high computational burden. In a complex scene with multiple small objects, the small objects belong to an identical category tend to have similar semantic co-occurrence information and simultaneously tend to have a similar aspect ratio, scale and appear in clusters in spatial layout. As such, human beings do not treat each region individually but integrate inter-object relationships, semantic or spatial, between regions. Such a phenomenon inspires us to explore how to model and infer the intrinsic semantic and spatial layout relationships for boosting small object detection.


To answer this question, we focus on recent works on modeling relationships and find that it is a common practice for introducing global contextual information into networks. For instance, PSP-Net \cite{zhao2017pyramid} and DenstASPP \cite{yang2018denseaspp} enlarge the receptive field of convolutional layers via combining multi-scale features to model the global relationships. Deformable CNN \cite{dai2017deformable} learns offsets for the convolution sampling locations, the scales or receptive field sizes can be adaptively determined. Moreover, Squeeze-and-Excitation Networks \cite{hu2018squeeze} (SE-Net) encodes the global information via a global average pooling operation to incorporate an image-level descriptor at every stage. However, these methods rely solely on convolutions in the coordinate space to implicitly model and communicate information between different regions. It is promising to squeeze out better performance if they can handle this problem effectively. On the contrary, Graph Convolutional Networks (GCN) is usually regarded as a composition of feature aggregation/propagation and feature transformation \cite{velivckovic2017graph}, thus enabling a global reasoning power that allows regions further away to directly communicate information with each other. As such, GCN is suitable for modeling and reasoning pair-wise high-order object relationships from the image itself which is expected to be helpful for boosting small object detection. 

In this paper, we propose a context reasoning approach based on GCN for small object detection to encode the implicit pair-wise regional relationships and propagate the semantic and spatial layout contextual information between regions. The flowchart of relationship construction is illustrated in \figref{fig:motivation} (b). It involves three modules: a semantic module for modeling the sparse semantic relationships from the initial regional features, a spatial layout module for modeling the sparse spatial layout relationships from the position and shape information of objects and a context reasoning module for integrating the sparse semantic and spatial layout contextual information to generate the dynamic scene graph and propagate the contextual information between objects. Experimental results show that the proposed approach can effectively boost the small object detection.


The contributions of this work are summarized as follows: 1) We propose a context reasoning approach that can effectively propagate the contextual information between regions and update the initial regional features for boosting the small object detection. 2) We design a semantic module and a spatial module for modeling the semantic and spatial layout relationships from the image itself without introducing external handcraft linguistic knowledge, respectively. Such relationships are beneficial for identifying small objects that fall into an identical category in the same scenario. 3) Comprehensive experiments are conducted and illustrate that our proposed approach can effectively boost the small object detection.

\section{Related Work}
\myPara{Object Detection.} Object detection is a fundamental problem in the computer vision field, and it is popularized by both two-stage and single-stage detectors. Two-stage detectors are developed from the R-CNN architecture \cite{girshick2014rich}, which firstly generates RoIs (Region of Interest) via some low-level computer vision algorithm \cite{zitnick2014edge,uijlings2013selective}, and then classify and locate them. The SPPNet \cite{he2015spatial} and Fast R-CNN \cite{girshick2015fast} exploit the spatial pyramid pooling to generate the shared feature once and then generate region feature from RoI pooling. In this manner, the redundant computation of feature extraction in R-CNN can be effectively reduced. Faster R-CNN \cite{ren2015faster} can further improve the effectiveness since it introduces a region proposal network (RPN) to replace the original stand alone time-consuming region proposal methods. For the sack of avoiding RoI-wise head work, R-FCN \cite{dai2016r} constructs position-sensitive score maps through a fully convolutional network. Moreover, the RoI Align layer proposed in Mask R-CNN \cite{he2017mask} can effectively address the coarse spatial quantization problem. FPN \cite{lin2017feature} integrates the low-resolution, semantically strong features with high-resolution, semantically weak features via a top-down pathway and lateral connections to address the scale variance. Conventionally, the two-stage detectors can achieve impressive performance but often at a high computational cost, make it hard to meet the requirements of real-time applications. To alleviate this dilemma, single-stage detectors avoid the time-consuming proposal generating step and classify the predefined anchors using CNNs directly, which are popularized by YOLO \cite{redmon2016you,redmon2017yolo9000} and SSD \cite{liu2016ssd}. RetinaNet \cite{lin2017focal} proposes Focal Loss to reduce the loss weight for easy samples, lead to a smaller performance gap between single-stage detectors and two-stage detectors.

However, existing object detectors suffer from a performance bottleneck in complex scenes with multiple small objects since it is hard for them to strike a balance between capturing semantically strong features and retaining more spatial information. Moreover, they treat each region individually and ignore the relationships between objects which leaves room for further exploration of their performance.

\myPara{Small Object Detection.} Small object detection is one of the common problems for the existing detection framework. In the field of tiny face detection, Bai \etal~\cite{bai2018finding} proposed to employ a super-resolution network to up-sample a blurry low-resolution image to fine-scale high-resolution one, which is in hope of supplementing the spatial information in advance. Later, in \cite{bai2018sod}, Bai \etal~proposed a multi-task generative adversarial network to recover detailed information for more accurate detection.

Regardless of their impressive performance, they suffer from a high computational burden since they introducing additional super-resolution network. They fail in mining the correlation between regions, which limits their small object detection performance improvements.

\begin{figure*}[t]
\begin{center}
   \includegraphics[width=1.0\linewidth]{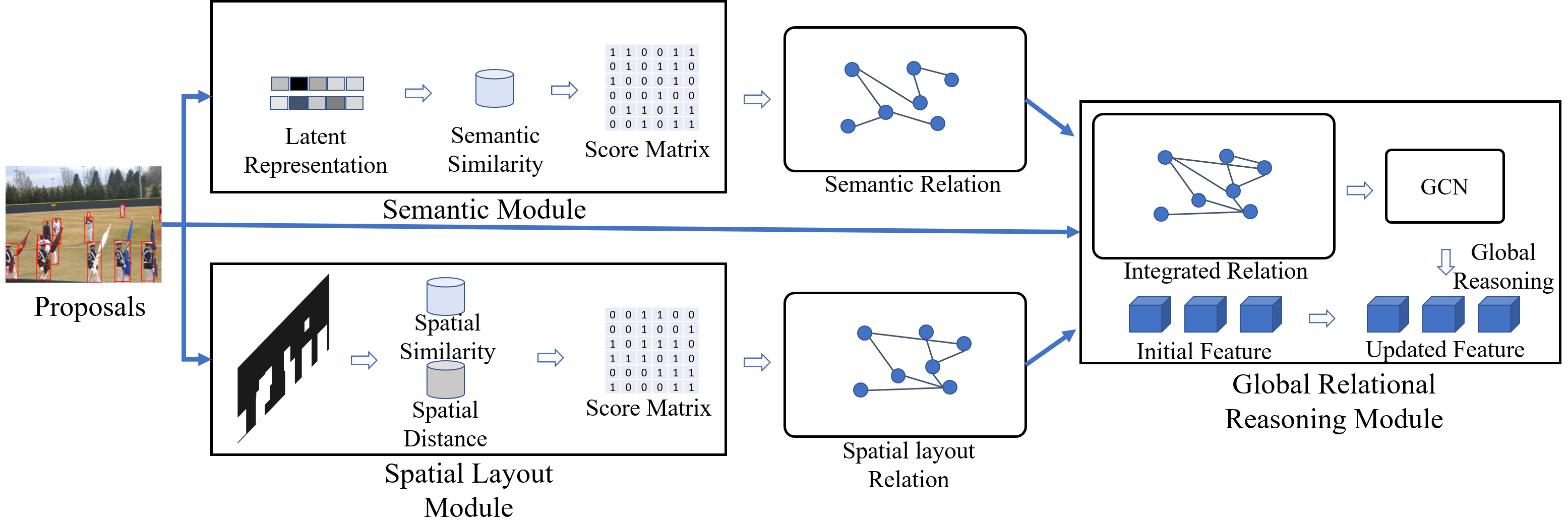}
\end{center}
   \caption{The overview of the proposed context reasoning framework. It consists of three module. A semantic module encodes the intrinsic semantic relationships from the initial regional features. A spatial layout module encodes the intrinsic spatial layout relationships from the position and shape information of objects. A context reasoning module integrates the contextual information between the objects and sparse relationships, and updates the initial regional features.}
\label{fig:framework}
\end{figure*}
\myPara{Relationship Mining.} Relationship mining aims to reasonable interacting, propagating and variating the information between objects and scenes. It has been applied in some common visual tasks, such as classification \cite{marino2016more}, object detection \cite{chen2018iterative} and visual relationship detection \cite{dai2017detecting}. A common practice in previous works \cite{akata2013label,almazan2014word,lampert2009learning,misra2017red} is to consider manual designed relationships and shared attributes among objects. For example, some works \cite{frome2013devise,mao2015learning,reed2016learning} try to reason via modeling the similarity such as the attributes in the linguistic space. The graph structure \cite{chen2018iterative,dai2017detecting,kipf2016semi,marino2016more} also demonstrates its amazing ability in incorporating external knowledge. In \cite{deng2014large}, Deng \etal~construct a relation graph from labels to guide the classification. Similarly, Chen \etal~\cite{chen2018iterative} design an iteratively reasoning framework that leverages both local region-based reasoning and global reasoning to facilitate object recognition.

However, these works rely on external handcraft linguistic knowledge, which requires laborious annotation work. Moreover, the handcraft knowledge graph usually is not so appreciated since the gap exists between linguistic and visual context. Some works \cite{hu2018relation,liu2018structure,norcliffe2018learning} propose to construct implicit relations from the image itself. Especially, Liu \etal~\cite{liu2018structure} encodes the relations via constructing a Structure Inference Network (SIN) which learns a fully-connected graph implicitly with stacked GRU cell. However, the redundant information and the inefficiency brought by a fully-connect graph make this method stagnant. We hope to imitate the human visual mechanism and construct a dynamic scene graph by mining the intrinsic semantic and spatial layout relationships from each image to facilitate small object detection.

\section{Proposed Approach}
In this section, we present our approach in detail. We first briefly overview the whole approach, and then expatiate on the semantic module and the spatial layout module, respectively. Finally, we present the details of a context reasoning module.
\subsection{Overview}

We start with an overview of the context reasoning framework before going into detail below. The system framework of our approach is shown in \figref{fig:framework}. Note that our context reasoning approach is flexible and can be easily injected into any two-stage detection pipelines. The human visual system tends to assign objects that have similar semantic co-occurrence information, aspect ratios, and scales to an identical category, which is beneficial for recognizing small objects in complex scenarios. Our approach mimics such a human visual mechanism and captures the inter-object relationships (both semantic and spatial layout) between small objects. It aims at inferring the existence of hard-to-detect small objects by measuring their relatedness to other easy-to-detect ones. In this paper, we explore whether mining the semantic and spatial layout relationships can boost small object detection.

We first construct a semantic module for encoding the intrinsic semantic relationships from the initial regional features and a spatial layout module for encoding the spatial layout relationships from the position and shape information of objects. Then both the semantic and the spatial layout relationships are fed into a context reasoning module and generate a region-to-region undirected graph $G=\langle\mc{N},\mc{E}\rangle$, where $\mc{N}$ are region nodes and each edge $e_{ij}\in \mc{E}$ encodes both semantic and spatial layout relationships between nodes. Finally, the context reasoning module integrates the contextual information between the objects and sparse relationships, which is further fused with the original regional features.

\subsection{Semantic Module}
\begin{figure}[t]
\begin{center}
   \includegraphics[width=1.\columnwidth]{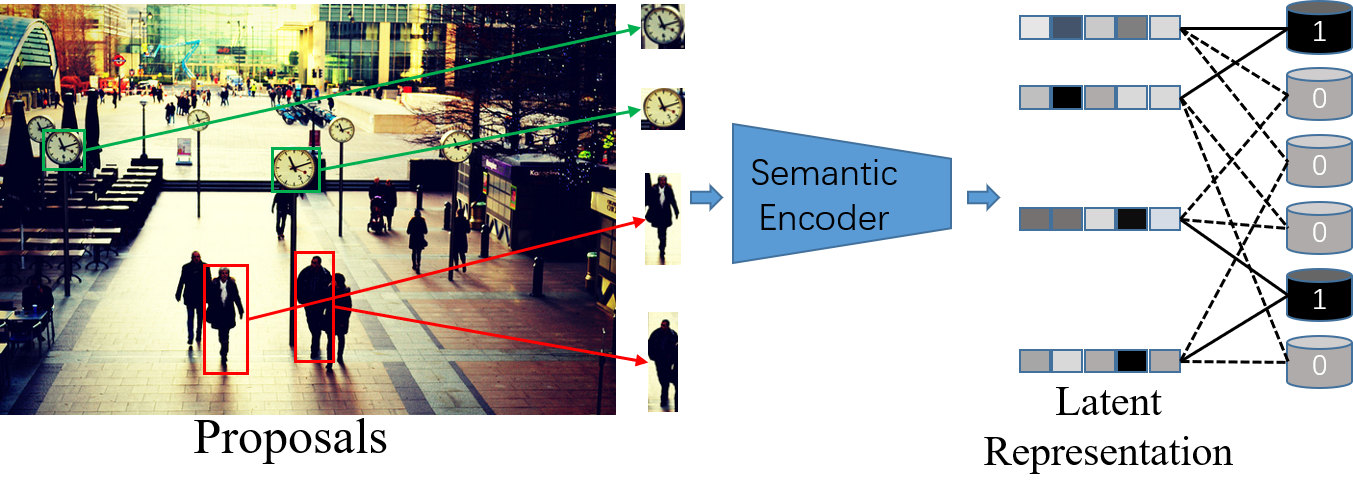}
\end{center}
   \caption{Flowchart of semantic relatedness calculation. The initial regional features from these proposals are fed into a semantic encoder to yield latent representations, which are used to calculating the relatedness from a learnable semantic relatedness function. Proposals fall into the same category tend to have similar semantic co-occurrence information lead to high relatedness and low if they not.}
\label{fig:senmanticExp}
\end{figure}
This module is learnable and aims to imitate the human visual mechanism to model the intrinsic semantic relationships between objects. As shown in \figref{fig:senmanticExp}, proposals fall into the identical category tend to have similar semantic co-occurrence information, lead to high relatedness and low if they not. More intuitively, a hard-to-detect small object, which has ambiguous semantic information, is more likely to be a clock if it has the top semantic similarities to some easy-to-detect clocks in the same scenario. The semantic context information of these easy-to-detect clocks tends to be beneficial for recognizing such a hard-to-detect object. We define a dynamic undirected graph $G_{sem}=\langle\mc{N},\mc{E}_{sem}\rangle$ to encode the semantic relationships from each image. Note that each node in $\mc{N}$ corresponding to a region proposal while each edge $e_{ij}^{'}\in\mc{E}_{sem}$ represents the relationship between nodes. Given $\mc{N}_r=|\mc{N}|$ proposal nodes, we first construct a fully-connect graph that contains $O(\mc{N}_r^2)$ possible edges between them. However, most of the connections are invalid due to regularities in real-world object interactions. A direct solution to this problem is to calculate the semantic relatedness between the fully-connected graph and then retain the relationships in high relatedness meanwhile prune the relationships in low relatedness. The flowchart of relatedness calculation is illustrated in \figref{fig:senmanticExp}.

Inspired from \cite{yang2018graph}, given initial regional feature pool $\mathbf{P}^o\in \mb{R}^{\mc{N}_r\times D}$, in which $D$ is the dimension of the initial regional features, we define a learnable semantic relatedness function $f(\cdot, \cdot)$ to calculate the semantic relatedness from each pair-wise initial regional features $\langle p_i^o,p_j^o\rangle\in \mathbf{P}^o$ in the original fully-connected graph. The semantic relatedness $s_{ij}^{'}$ can be formulated as
\begin{equation}
s_{ij}^{'}=\delta(i,j)\cdot f(p_i^o,p_j^o)=\delta(i,j)\cdot \Phi(p_i^o)\Phi(p_j^o)^{T},
\end{equation}
where $\delta(i,j)$ is an indicator function that equals $0$ if the $i$th and $j$th regions are highly overlapped with each other and $1$ otherwise. $\Phi(\cdot)$ is a projection function that projects the initial regional features to latent representations. Since different regions are parallel and there is no subject and object division, we set it to a multi-layer perceptron (MLP) to encode undirected relationships in this paper. A sigmoid function is applied to the score matrix $S^{'}=\{s_{ij}^{'}\}$ for normalizing all the scores range from $0$ to $1$. Then we sort the score matrix $S^{'}$ by rows and preserve the top K values in each row. The pair-wise regional relationships corresponding to the preserved values are set as the selected relationships. The value of adjacent edge $e_{ij}^{'}$ is set to $1$ if the corresponding region-to-region relationship is selected and $0$ otherwise.

The semantic module maps the original region feature that involves rich semantic and location information into a new feature space via an MLP architecture and preserves the regions with the high similarity of corresponding features. In the training process, the location information tends to be ignored and the semantic information tends to be preserved since the high similarity of location information will result in retaining regions with a high overlap ratio and such regions will be suppressed by NMS algorism. Thus, it encodes the semantic information. In this manner, we can obtain a sparse semantic relationships $\mc{E}_{sem}$ that most informative edges are retained and the noising edges are pruned.

\subsection{Spatial Layout Module}
\begin{figure}[t]
\begin{center}
   \includegraphics[width=1.0\columnwidth]{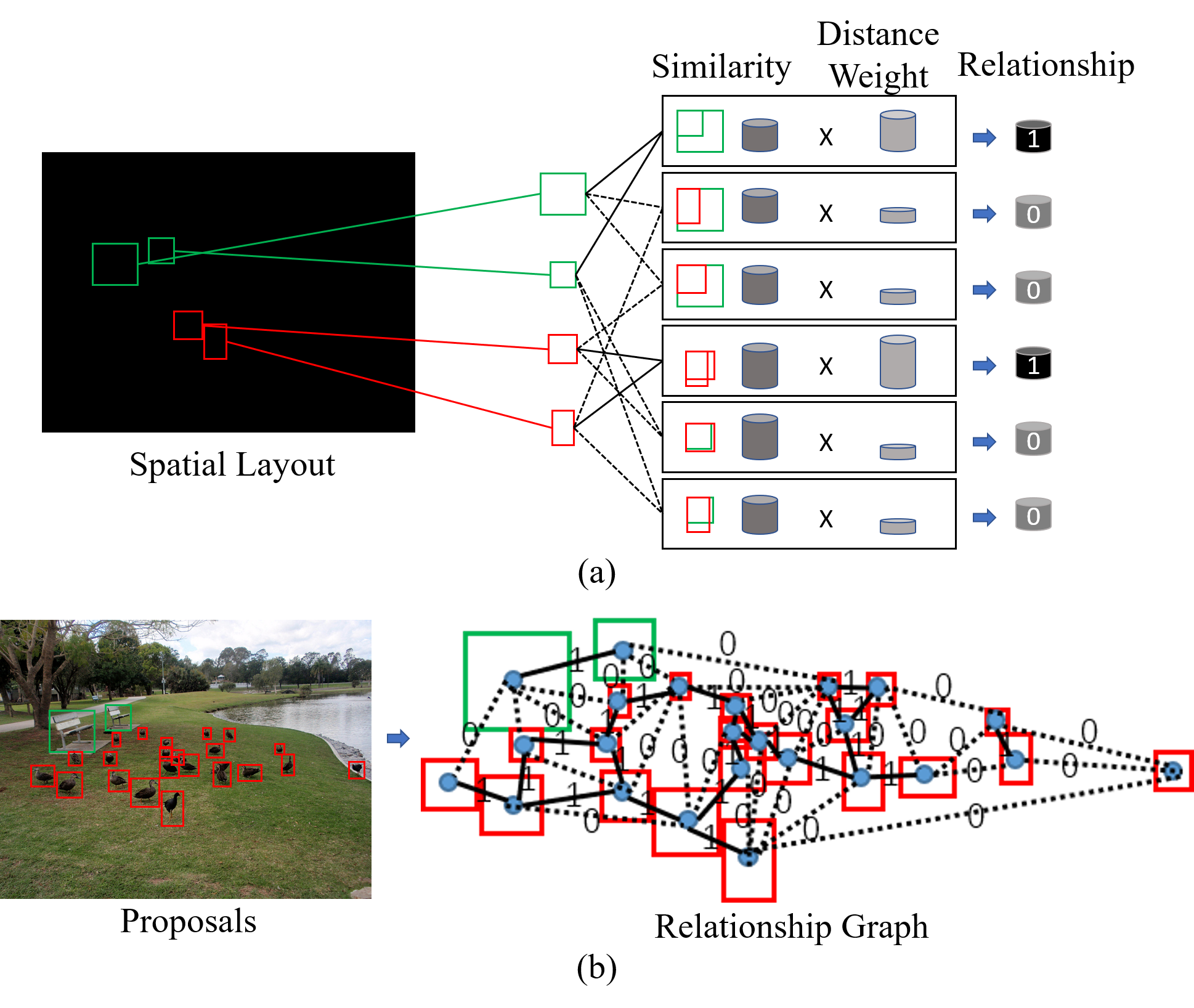}
\end{center}
   \caption{(a) Flowchart of spatial layout relatedness calculation. The spatial layout of each pair-wise region is fed into the spatial layout module to compute the spatial similarity and spatial distance weight for calculating the spatial layout relatedness. (b) An example of a spatial layout relationship graph.}
\label{fig:spatialExp}
\end{figure}

Conventionally, the small objects fall into the identical category in the scene tend to have similar spatial aspect ratios and scales, for instance, the two chairs in \figref{fig:spatialExp} (b) are in a high spatial similarity but not so between chairs and the majority birds. Meanwhile, this is not a one-size-fits-all rule and we can easily find some failure cases in \figref{fig:spatialExp} (b), a few birds are in high spatial similarity with the chairs but in different categories. This suggests that we should revisit the question of how to effectively model the spatial layout relationships between small objects for better recognition. We can find that the chairs are closer to each other than they are to most birds, and the birds are in a similar situation. This phenomenon can be generalized to the majority of scenarios, that is, small objects of the identical category tend to appear in clusters in spatial layout. Inspired by this, we construct the spatial layout module to model the intrinsic spatial layout relationships from both spatial similarity and spatial distance. Its flowchart is as shown in \figref{fig:spatialExp} (a).

We define a spatial layout dynamic undirected graph $G_{spa}=\langle\mc{N},\mc{E}_{spa}\rangle$ to encode the spatial layout relationships. Similar to that in the semantic module, we define a spatial layout relatedness function $g(\cdot, \cdot)$ to calculate the relatedness in the original fully-connected graph. The spatial layout relatedness $s_{ij}^{''}\in S^{''}$ can be formulated as
\begin{equation}
s_{ij}^{''}=\delta(i,j)\cdot g(C_i^o,C_j^o)=\delta(i,j)\cdot m_{ij}^{r}\cdot w_{ij}^r,
\end{equation}
where $C_i^o=(x_i,y_i,w_i,h_i)$ and $C_j^o=(x_j,y_j,w_j,h_j)$ are region coordinates corresponding to region $i$ and $j$, respectively. $m_{ij}^{r}$ and $w_{ij}^r$ are spatial similarity and spatial distance weight, respectively.
\begin{eqnarray}
&m_{ij}^{r}=\frac{\min(w_i,w_j)\cdot\min(h_i,h_j)}{w_ih_i + w_jh_j - \min(w_i,w_j)\cdot\min(h_i,h_j)}, \\
&w_{ij}^r=\exp(-\lambda \cdot m_{ij}^{d}),
\end{eqnarray}
where $\lambda$ is functioned as a scale parameter which is empirically set to $5e-4$ in this paper. $m_{ij}^{d}$ is the spatial distance between the centers of the two regions. We sort the score matrix $S^{''}$ by rows and preserve the top K values in each row. The pair-wise regional relationships corresponding to the preserved values are set as the selected relationships. Finally, we set the adjacent edge $e_{ij}^{''}\in\mc{E}_{spa}$ in the same manner as in semantic module. A constructed spatial layout graph is illustrated in \figref{fig:spatialExp} (b).
\subsection{Context Reasoning Module}
The context reasoning module is constructed to integrate the contextual information between the objects and sparse relationships. Given the initial regional features $\mathbf{f}\in\mb{R}^{\mc{N}_r\times D}$ and the encoded semantic and spatial layout relationships, we need to select the relationships that are highly related to each other, semantic or spatial layout. We fuse the semantic and spatial layout relationships via
\begin{equation}
\mc{E}=\mc{E}_{sem}\cup \mc{E}_{spa}.
\end{equation}

\begin{figure}[t]
\begin{center}
   \includegraphics[width=1.0\columnwidth]{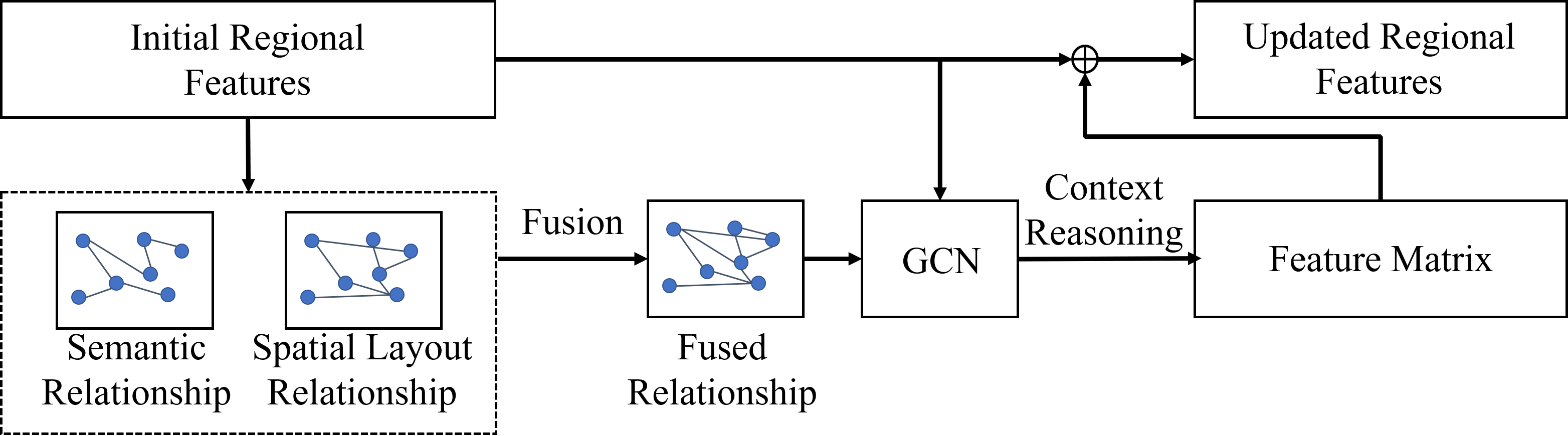}
\end{center}
   \caption{The context reasoning flowchart. The semantic and spatial layout relationships are fused for propagating both the semantic and spatial layout contextual information via a GCN. The original regional features are updated with the output of the GCN.}
\label{fig:reasoning}
\end{figure}

The connections between regions are non-Euclidean data and high irregular, which can not be systematically and reliably processed by CNNs in general. Graph Convolutional Network (GCN) is capable for better estimating edge strengths between the vertices of the fused relationship graph $\mc{E}$, thus leading to more accurate connections between individuals. Intuitively, information communication between regions with high relatedness is capable provide more effective contextual information, which will effectively boost small object detection. As a result, we construct a light-weight GCN for regional context reasoning. Its flowchart is illustrated in \figref{fig:reasoning}. It consists of $L>0$ layers each with the same propagation rule defined as follows. We define $\mc{H}^{(l)}\in\mb{R}^{\mc{N}_r\times D}$ as the hidden feature matrix of the $l$-th layer and $\mc{H}^{(0)}=\mathbf{f}$. The $\mc{H}^{(l)}$ can be formulated as
\begin{equation}
\mc{H}^{(l)}=\sigma(\mc{D}^{-\frac{1}{2}}\tilde{\mc{E}}\mc{D}^{-\frac{1}{2}}\mc{H}^{(l-1)}\mc{W}^{l}),
\end{equation}
where $\mc{D}$ is the degree matrix of $\mc{E}$ while $\tilde{\mc{E}}=\mc{D}-\mc{E}$ is a combinatorial laplacian matrix of $G$. $\mc{W}^{l}$ denotes the trainable weight matrix of the $l$-th layer, and $\sigma(\cdot)$ is LeakyReLU activation function. The initial regional features $\mathbf{f}$ are updated with the output of GCN
\begin{equation}
\tilde{\mathbf{f}}=\mathbf{f}\oplus \mc{H}^{L},
\end{equation}
where $\tilde{\mathbf{f}}$ and $\oplus$ represent the updated features and element-wise addition operation, respectively.

In this manner, both co-occurrence semantic and spatial layout information can effectively propagate to each other, which enables the model a better self-correction ability compared with before, and the problems of false and omissive detection are alleviated.

\begin{table*}[t]
\footnotesize
\centering
\begin{threeparttable}
\caption{Comparison with state-of-the-art detectors on COCO \textit{test-dev}. We show results for our IR R-CNN with backbone ResNet-50 and ResNet-101. Our module achieves top results in small object detection, outperforming most one-stage and two-stage models. The best, runner-up and second runner-up two-stage models are marked with {\color{red}red}, {\color{green}green} and {\color{blue}blue}, respectively.}
\label{tab:performanceCom}
\setlength{\tabcolsep}{4.3mm}{
\begin{tabular}{c|l|c|cccccc}
\toprule
\multicolumn{2}{c|}{} &backbone   &$AP$   &$AP_{50}$  &$AP_{75}$  &$AP_{S}$   &$AP_{M}$   &$AP_{L}$   \tabularnewline
\midrule
\multirow{7}{*}{\begin{sideways}\bl{one-stage}\end{sideways}}&YOLOv2~\cite{redmon2016you}   &DarkNet-19 &21.6   &44.0   &19.2   &5.0   &22.4   &35.5 \tabularnewline
&SSD513~\cite{fu2017dssd}                 &ResNet-101             &31.2   &50.4   &33.3   &10.2   &34.5   &49.8 \tabularnewline
&YOLOv3~\cite{redmon2018yolov3}                 &Darknet-53             &33.0   &57.9   &34.4   &18.3   &35.4   &41.9 \tabularnewline
&DSSD513~\cite{fu2017dssd}                &ResNet-101             &33.2   &53.3   &35.2   &13.0   &35.4   &51.1 \tabularnewline
&RefineDet512~\cite{zhang2018single}           &ResNet-101             &36.4   &57.5   &39.5   &16.6   &39.9   &51.4 \tabularnewline
&RetinaNet~\cite{lin2017focal}              &ResNet-101             &39.1   &59.1   &42.3   &21.8   &42.7   &50.2 \tabularnewline
&CornerNet511~\cite{law2018cornernet}\tnote{*}           &Hourglass-104          &40.5   &56.5   &43.1   &19.4   &42.7   &53.9 \tabularnewline
\midrule
\multirow{9}{*}{\begin{sideways}\bl{two-stage}\end{sideways}}&Faster R-CNN+++~\cite{he2016deep}\tnote{*}  &ResNet-101 &34.9   &55.7   &37.4   &15.6   &38.7   &50.9 \tabularnewline
&Faster R-CNN by G-RMI~\cite{huang2017speed}  &Inc-ResNet-v2    &34.7   &55.5   &36.7   &13.5   &38.1   &52.0 \tabularnewline
&Faster R-CNN w FPN~\cite{lin2017feature}     &ResNet-101             &36.2   &59.1   &39.0   &18.2   &39.0   &48.2 \tabularnewline
&Faster R-CNN w TDM~\cite{shrivastava2016beyond}     &Inc-ResNet-v2    &36.8   &57.7   &39.2   &16.2   &39.8   &52.1 \tabularnewline
&Deformable R-FCN~\cite{dai2017deformable}\tnote{*}       &Aligned-Inc-ResNet    &37.5   &58.0   &40.8   &19.4   &40.1   &{\color{blue}52.5} \tabularnewline
&Mask R-CNN~\cite{he2017mask}             &ResNet-101             &38.2   &60.3   &41.7   &20.1   &41.1   &50.2 \tabularnewline
&Regionlets~\cite{hu2018relation}             &ResNet-101             &39.3   &59.8   &--     &21.7   &{\color{blue}43.7}   &50.9 \tabularnewline
&Fitness NMS~\cite{tychsen2018improving}   &ResNet-101             &{\color{green}41.8}   &{\color{blue}60.9}   &{\color{green}44.9}   &21.5   &{\color{green}45.0}   &{\color{red}{57.5}} \tabularnewline
&FRCNN-FD-WT~\cite{peng2019pod}   &ResNet-101             &{\color{red}{42.1}}   &{\color{red}{63.4}}   &{\color{red}{45.7}}   &{\color{blue}21.8}   &{\color{red}{45.1}}   &{\color{green}57.1} \tabularnewline
\midrule
&\bl{IR R-CNN}   &ResNet-50             &37.6   &60.0   &40.6   &{\color{green}21.9}   &39.7   &47.0 \tabularnewline
&\bl{IR R-CNN}   &ResNet-101             &{\color{blue}39.7}   &{\color{green}62.0}   &{\color{blue}43.2}   &{\color{red}{22.9}}   &42.4   &50.2 \tabularnewline
\bottomrule
\end{tabular}
}
\begin{tablenotes}
    \footnotesize
    \item[*] Models used bells and whistles at inference.
\end{tablenotes}
\end{threeparttable}
\end{table*}
\section{Experiments}
In this section, experiments are conducted to evaluate the effectiveness of our proposed approach. We will begin with our experimental settings and then present the implementation details and benchmark the state-of-the-art models, finally, we present a detailed performance analysis.
\subsection{Experimental Settings}

We evaluate our proposed approach on the bounding box detection track of the challenging COCO benchmark \cite{lin2014microsoft}, which has more small objects than large/medium objects, approximately $41\%$ of objects are small (area$<32^2$). With respect to prior investigation of \cite{bell2016inside,lin2017feature}, we train the COCO \textit{trainval135k} split (union of $80$k train images and random $35$k subset of val images). We report the ablation studies by evaluating the \textit{minival} split (the remaining $5$k images from val images). For a fair comparison, we report the performance on \textit{test-dev} split, which has no public labels and requires the use of the evaluation server.

According to the scale of objects, the COCO dataset can be divided into three subsets: small, medium and large. In detail, the large objects with an area larger than $96^2$, the small objects with an area smaller than $32^2$, the medium objects with an area in between. In this paper, we focus on the performance of small object detection. The standard COCO metrics are reported in this paper, including $AP$ (averaged over IoU thresholds), $AP_{50}$, $AP_{75}$, and $AP_{S}$, $AP_{M}$, $AP_{L}$ (AP at different scales).

\subsection{Implementation Details}
We re-implement Faster R-CNN \cite{ren2015faster}, with ResNet-50 and ResNet-101 as backbones, as our baseline methods in PyTorch \cite{paszke2017automatic}. Note that our network backbone is pre-trained on ImageNet \cite{russakovsky2015imagenet} and then fine-tuned on the detection dataset. The parameters in MLP architecture and context reasoning module are randomly initialized and are trained from scratch. The overall network is trained in an end-to-end manner, and its input images are resized to have a short side of $800$ pixels. It is trained with stochastic gradient descent (SGD). We use synchronized SGD over $4$ GPUs with a total of $16$ images per minibatch (4 images per GPU). The model is trained for $90$k iterations with an initial learning rate of $0.02$. We decay the learning rate at $60$k and again at $80$k iterations with decay rate $0.1$. We
use a weight decay of $0.0001$ and momentum of $0.9$. We empirically set $K=64$ in the relationship graph construction $L=2$ in the context reasoning module, respectively.

\begin{figure*}[t]
\begin{center}
   \includegraphics[width=1.\textwidth]{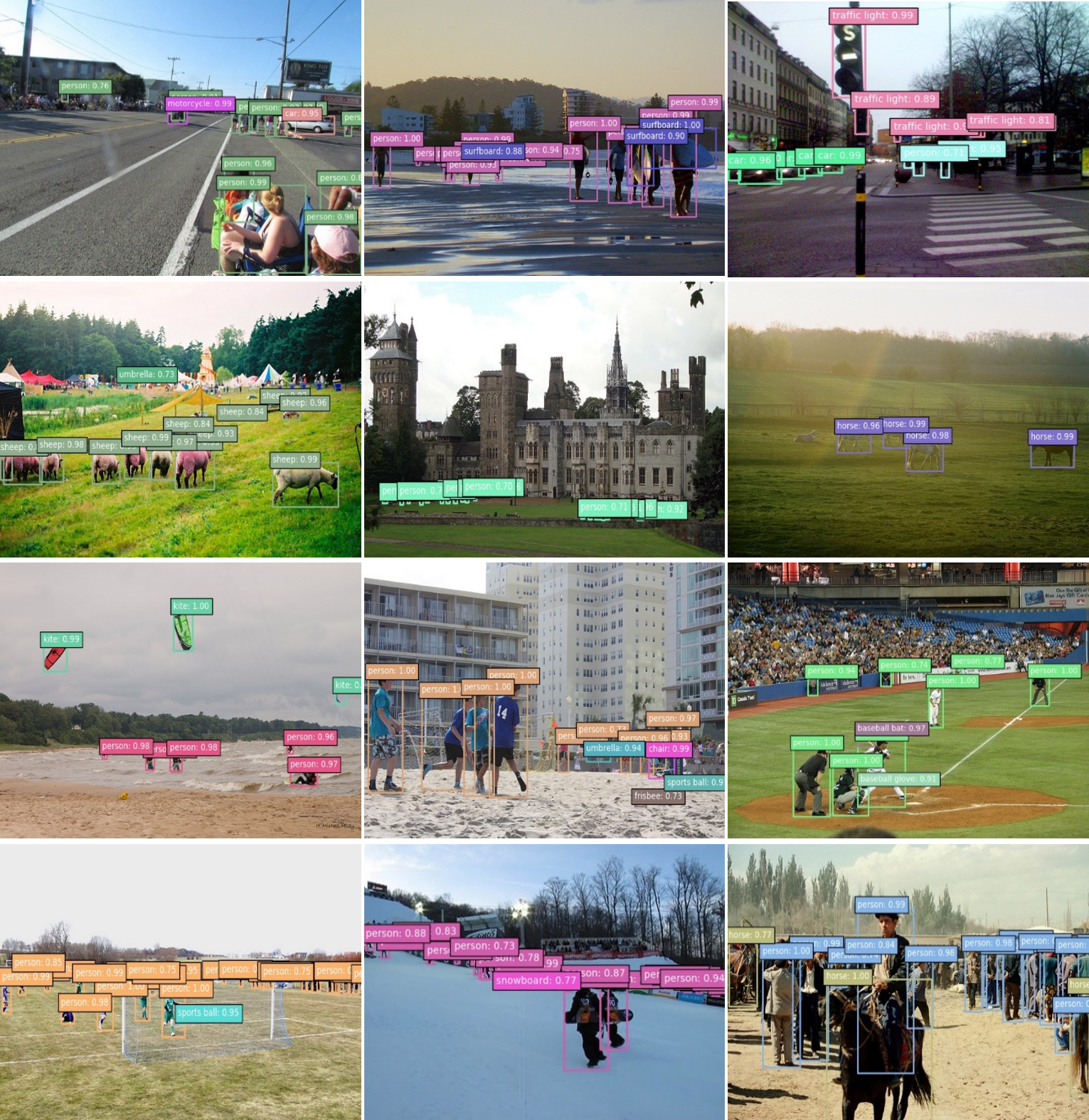}
\end{center}
   \caption{Qualitative results of our IR R-CNN with ResNet-101 as backbone. The model is trained on COCO \textit{trainval135k} split (union of $80$k train images and random $35$k subset of val images).}
\label{fig:representative}
\end{figure*}

\subsection{Comparison with the State-of-the-art Models}

We evaluate our proposed approach to the bounding box detection task of the challenging test COCO dataset. We compare it with several state-of-the-art models, including both one-stage and two-stage models, and their performance is as shown in \tabref{tab:performanceCom}. From this table, we find that our proposed approach can achieve better accuracy than the popular models in small object detection.
This reveals that our approach can strongly improve the original small object regional features, and the correctness of the theory that modeling the semantic and spatial layout relationships to boost the small object detection with only a 6.9\% parameter increment (60.6 million$\rightarrow$64.8 million parameters). Note that our approach is designed for the complex scenes with multiple small objects, make it flexible and portable for diverse detection systems to improve the small object detection performance.
Some qualitative examples of detection results generated by our IR R-CNN are illustrated in \figref{fig:representative}. We observe that our approach can detect the most objects that conform to the human visual cognitive system, even if there are very small objects in the scene. This indicates the effectiveness of our approach in modeling the relationships between small objects, semantic and spatial layout. However, we can also find some failure cases, which shows that our method still has room for improvements to promote the performance of small object detection.

\subsection{Detailed Performance Analysis}
We conduct several experiments on COCO \textit{minival} to verify the effectiveness of the proposed approach. Unless otherwise stated, all models in detailed performance analysis are implemented on Faster R-CNN with ResNet-50 as the backbone.

\myPara{Parameter Analysis.} We conduct an experiment to evaluate the parameter K in \{16, 32, 64, 96\}. The performance of the proposed approach with different K is summarized in \tabref{tab:parameterAnalysis}. From this table, we find that the overall detection performance remains relatively stable, while the performance of small object detection improves substantially as K grows and it peaks at K=64. However, when the K continues to grow, the performance of small object detection decays.

This can be interpreted as that low K will result in the proposed semantic and spatial layout module that can not encode sufficient semantic and spatial layout relationships, respectively. This constricts the semantic and spatial layout context information that can be propagated between regions and leads to inferior small object detection performance. On the contrary, large K increases the risk of unnecessary relationships being encoded. In other words, noise may be introduced, which has a negative impact on the improvements of small object detection. In summary, the performance improvements can be maximized when the appropriate K enables sufficient relationships to be encoded and effectively propagates context information between regions while avoiding the introduction of noise.


\begin{table}[t]
\footnotesize
\centering
\begin{threeparttable}
\caption{Parameter analysis on \textit{minival} subset.}
\label{tab:parameterAnalysis}
\setlength{\tabcolsep}{3.2mm}{
\begin{tabular}{c|cccccc}
\toprule
K            &$AP$       &$AP_{50}$  &$AP_{75}$  &$AP_{S}$   &$AP_{M}$   &$AP_{L}$   \tabularnewline
\midrule
16           &36.7      &58.6      &39.7      &21.3      &40.0      &47.5      \tabularnewline
32           &37.2      &59.4      &39.8      &22.1      &40.1      &48.0      \tabularnewline
64           &37.3      &\bl{59.5} &\bl{40.5} &\bl{22.9} &40.5      &\bl{48.5} \tabularnewline
96           &\bl{37.4} &59.5      &40.5      &22.0      &\bl{40.6} &48.4      \tabularnewline
\bottomrule
\end{tabular}
}
\end{threeparttable}
\end{table}


\begin{table}[t]
\footnotesize
\centering
\caption{Ablation study on \textit{minival} subset.}
\label{tab:performanceRelation}
\setlength{\tabcolsep}{2.4mm}{
\begin{tabular}{cc|cccccc}
\toprule
Sem               &Spa          &$AP$       &$AP_{50}$  &$AP_{75}$  &$AP_{S}$   &$AP_{M}$   &$AP_{L}$   \tabularnewline
\midrule
{}                &{}           &36.8      &58.7      &39.6      &21.0      &39.9      &47.7      \tabularnewline
\checkmark        &{}           &36.6      &58.5      &39.6      &22.3      &40.0      &47.3      \tabularnewline
{}                &\checkmark   &37.0      &59.0      &40.2      &21.9      &40.2      &47.8      \tabularnewline
\checkmark        &\checkmark   &\bl{37.3} &\bl{59.5} &\bl{40.5} &\bl{22.9} &\bl{40.5} &\bl{48.5} \tabularnewline
\bottomrule
\end{tabular}
}
\end{table}

\myPara{Ablation Studies.} Ablation studies, which mainly consists of two different settings, are conducted to verify the effectiveness of the proposed semantic and spatial layout modules. In the first setting, we only consider the semantic relationships and ignore the spatial layout relationships for context reasoning. In this manner, only the regions in high semantic similarity are propagating context information with each other. In the second setting, similarly, we ignore the semantic relationships between regions and only fed the spatial layout relationships into the context reasoning module for further reasoning. Tab. \ref{tab:performanceRelation} summarizes the performance of ablation studies on \textit{minival} subset.


From this table, we find that both the semantic and spatial layout module can boost the small object detection to some extent. But their respective improvements are quite limited when compared to the full model. This can be interpreted as the semantic module that is capable to encode semantic relations from semantic similarity, enable the context reasoning module to propagate the high-order semantic co-occurrence contextual information between objects, which leads to a performance gain. However, it is not so beneficial for small objects that are hard to extract semantically strong features but fall into the identical category. The spatial layout module sets aside the semantic similarity and constructs relations from spatial layout, gives the small objects, that in high spatial similarity and appear in clusters in spatial layout, an opportunity to propagate spatial layout contextual information to each other. This can alleviate the problems in the semantic module but in high risk to introducing noise. Since the two modules can complement to each other, the fusion of them naturally enables the performance gains maximum. Specially, \tabref{tab:performanceRelation} reveals that our context reasoning approach can boost the performance of small object detection by 1.9 points on \textit{minival} subset.

\section{Conclusion}
We present a novel context reasoning approach for small object detection which models and infers the intrinsic semantic and spatial layout relationships between objects. It constructs sparse semantic relationships from the semantic similarity and sparse spatial layout relationships from the spatial similarity and spatial distance. A context reasoning module takes the semantic and spatial layout relations as input, propagates the semantic and spatial layout contextual information for updating the initial regional features, which make it capable for the object detectors to alleviate the problem of false and omissive detection for small objects. The experimental results on COCO have validated the effectiveness of the proposed approach. We believe that the IR R-CNN could benefit the current small object detection with relationship modeling and inference.

In future work, we will tentatively explore the feasibility of introducing orientation information into the context reasoning module, as well as combing both intrinsic relationship and external handcraft linguistic knowledge for further small object detection performance improvements.

\bibliographystyle{ACM-Reference-Format}
\bibliography{irebib}

\appendix

%
%
%
%
%
%
%

\end{document}